\documentclass[letterpaper]{article} 
\usepackage{aaai2026}  
\usepackage{times}  
\usepackage{helvet}  
\usepackage{courier}  
\usepackage[hyphens]{url}  
\usepackage{graphicx} 
\usepackage{natbib}  
\usepackage{caption} 
\usepackage{algorithm}
\usepackage{algorithmic}
\usepackage{amsmath}
\usepackage{color, soul}
\usepackage{multirow}
\usepackage[table,xcdraw]{xcolor}
\usepackage{amssymb}
\usepackage{subcaption}
\usepackage{arydshln}
\nocopyright

\usepackage{newfloat}
\usepackage{listings}
\DeclareCaptionStyle{ruled}{labelfont=normalfont,labelsep=colon,strut=off} 
\floatstyle{ruled}
\newfloat{listing}{tb}{lst}{}
\floatname{listing}{Listing}
\title{Time-step Mixup for Efficient Spiking Knowledge Transfer \\ from Appearance to Event Domain}
\author{
Yuqi Xie\textsuperscript{1}\thanks{Equal contribution.} \quad
Shuhan Ye\textsuperscript{1,2}\footnotemark[1] \quad
Yi Yu\textsuperscript{2}\thanks{Corresponding authors: wangchong@nbu.edu.cn, yu.yi@ntu.edu.sg} \quad
Chong Wang\textsuperscript{1,3}\footnotemark[2] \quad
Qixin Zhang\textsuperscript{2} \\
Jiazhen Xu\textsuperscript{1} \quad
Le Shen\textsuperscript{1} \quad
Yuanbin Qian\textsuperscript{1} \quad
Jiangbo Qian\textsuperscript{1,3} \quad
Guoqi Li\textsuperscript{4}\\
}

\affiliations{

\textsuperscript{1}Ningbo University \quad
\textsuperscript{2}Nanyang Technological University\\
\textsuperscript{3}Merchants' Guild Economics and Cultural \quad
\textsuperscript{4}Institute of Automation, Chinese Academy of Sciences\\
2411100305@nbu.edu.cn, shuhan006@e.ntu.edu.sg, yu.yi@ntu.edu.sg, wangchong@nbu.edu.cn, qixin.zhang@ntu.edu.sg, \{2311100314, 2411100289, 2311100301, qianjiangbo\}@nbu.edu.cn,
guoqi.li@ia.ac.cn
}

\usepackage{bibentry}

\begin{document}

\maketitle

\begin{abstract}
The integration of event cameras and spiking neural networks holds great promise for energy-efficient visual processing. However, the limited availability of event data and the sparse nature of DVS outputs pose challenges for effective training. 
Although some prior work has attempted to transfer semantic knowledge from RGB datasets to DVS, they often overlook the significant distribution gap between the two modalities. 
In this paper, we propose Time-step Mixup knowledge transfer (TMKT), a novel fine-grained mixing strategy that exploits the asynchronous nature of SNNs by interpolating RGB and DVS inputs at various time-steps. To enable label mixing in cross-modal scenarios, we further introduce modality-aware auxiliary learning objectives. These objectives support the time-step mixup process and enhance the model's ability to discriminate effectively across different modalities. Our approach enables smoother knowledge transfer, alleviates modality shift during training, and achieves superior performance in spiking image classification tasks. Extensive experiments demonstrate the effectiveness of our method across multiple datasets. The code will be released after the double-blind review process.
\end{abstract}

\section{Introduction}
In recent years, the integration of event cameras and spiking neural networks (SNNs) has attracted significant attention. Event cameras, also known as dynamic vision sensors (DVS), are inspired by the mammalian brain and capture visual data in response to changes in light intensity. This makes them an ideal solution for addressing the limitations of conventional cameras, such as low dynamic range and frame rate \cite{atis, celex, davis}. Meanwhile, SNNs are inherently well-suited for processing event-driven inputs while offering impressive energy efficiency. Their ability to process temporal information aligns perfectly with the high temporal resolution provided by event cameras \cite{deng2021}. The synergy between these two bio-inspired technologies presents a compelling approach for tackling low-power vision tasks like image classification \cite{spikformer,ckd}, action recognition \cite{sdv3}, and video anomaly detection \cite{ucfcrimedvs}.

\begin{figure}[t]
\centering
\includegraphics[width=0.95\columnwidth]{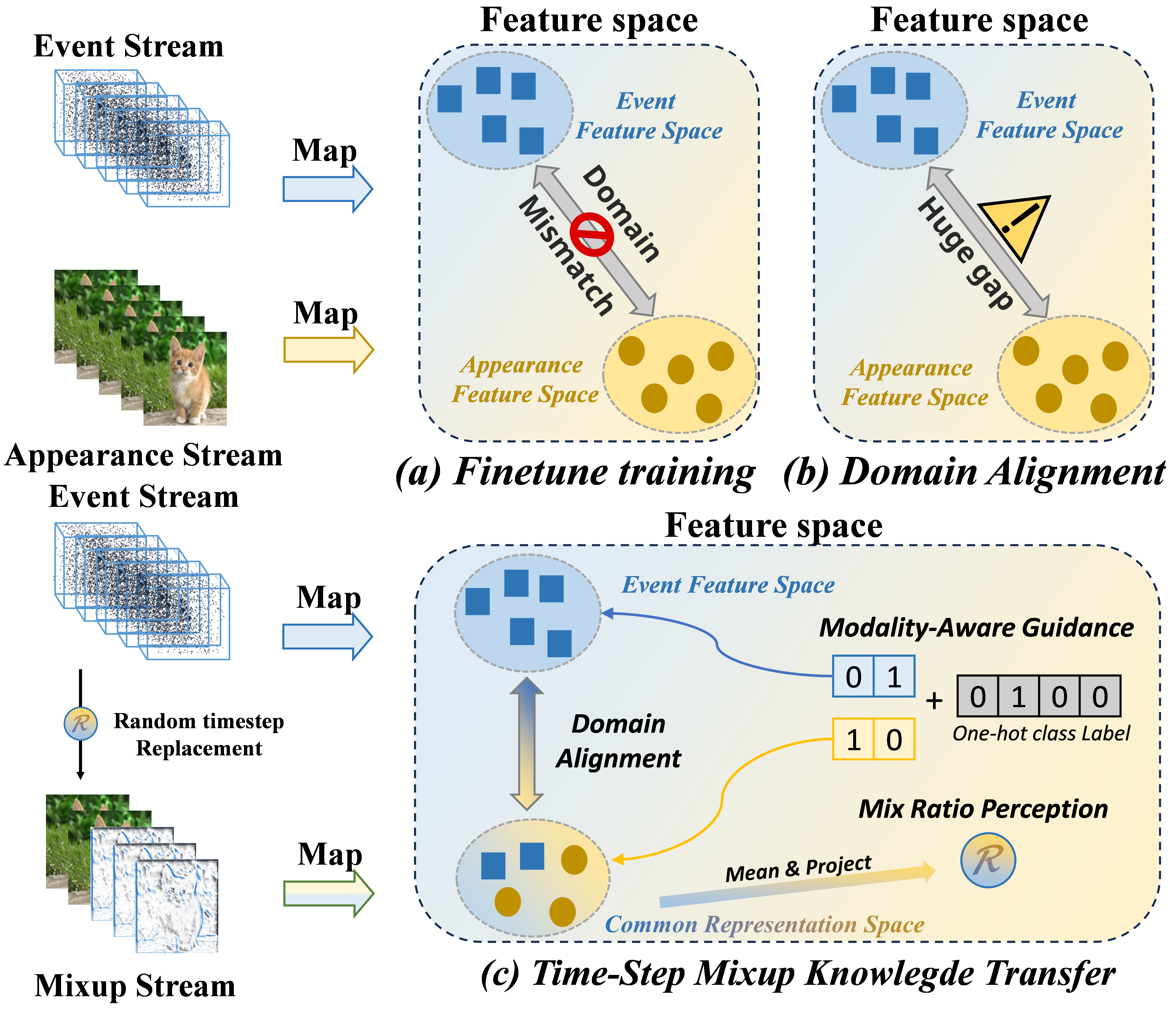} 
\caption{Different strategies for leveraging appearance data to assist spiking neural networks in learning the event domain. (a) Finetune training suffers from domain mismatch issues. (b) Domain alignment helps mitigate this issue to some extent. (c) Our Time-step Mixup offers a smoother learning paradigm, alleviating the convergence difficulties that arise during domain transition.}
\label{fig1}
\end{figure}


Despite their advantages, the application of event cameras faces significant challenges, due to the costly and time-consuming data acquisition process. This results in small-scale and hard-to-access datasets that hinder its further development. Additionally, since DVS only capture and encode brightness changes exceeding a certain threshold, they are primarily sensitive to the edges of moving objects. This means they discard a substantial amount of contextual cues, such as background, texture and color, which are crucial for comprehensive semantic understanding needed in high-level vision tasks.
In contrast, appearance datasets (RGB ones) offer abundant contextual information, and are both large-scale and easily accessible. 
These datasets provide richer details that support more robust analysis across various applications.

However, due to the significant distribution gap between appearance and event domains, as shown in Fig. \ref{fig1}, directly fine-tuning models pre-trained on appearance datasets often results in negative transfer. 
To address this issue, several studies have explored knowledge transfer methods aimed at bridging this modality gap by transferring rich semantic representations from the appearance domain to the event domain \cite{stl, ekt}. Despite these efforts, existing approaches still tend to overlook their intrinsic distributional difference at the raw data level. Specifically, RGB frames typically exhibit dense and intensity-rich pixel distributions, whereas DVS event frames are highly sparse with most values concentrated around zero and significantly different dynamic ranges.
The recent Knowledge-Transfer method \cite{ekt} addresses the challenge of data distribution by introducing a sliding data replacement strategy. In this approach, samples from the source domain (appearance) are gradually replaced with those from the target domain (event) during training. 
This allows the model to learn shared representations across both domains, progressively shifting from being source-dominated to target-dominated through a controlled ratio.
Nevertheless, this process operates under the implicit assumption that appearance and event data share similar feature distributions, which does not hold in practice.
As a result, when dissimilar modalities coexist within the same batch, it can lead to intra-batch modality shifts that complicate learning.

Inspired by the concept of data replacement, we aim to develop a smoother and more refined approach for cross-modal data mixing that guides the model in effectively handling heterogeneous modalities. 
A natural starting point is to examine successful MixUp-like strategies \cite{mixup,cutmix}, which provide valuable insights through their approach of spatially blending inputs and labels simultaneously. However, these techniques are primarily designed for intra-class interpolation within a single modality. The substantial gap between RGB and DVS data presents challenges that make direct input-level and label interpolation impractical in our scenario. 


To address these challenges, we propose Time-step Mixup Knowledge Transfer (TMKT), a novel knowledge transfer framework tailored for SNNs. 
Leveraging the inherent heterogeneity of SNNs, TMKT performs a Time-step Mixup strategy to mix and embed source and target domain data into a robust common representation space. 
This facilitates smoother and more gradual knowledge transfer, as illustrated in Fig.~\ref{fig1}(c). 
Furthermore, to enable the label interpolation in cross-modal settings and enhance the model’s awareness of modality-specific information, we introduce a novel modality-aware guidance label. 
Specifically, each input frame is augmented with a modality-aware indicator at the time-step level. 
This allows the network to distinguish between sources of different modality during training and adapt accordingly. 
The same indicator is also employed for perceiving multimodal mixup ratio, ensuring smoother learning transitions. 
Combining these, our method achieves strong performance on image classification tasks, demonstrating the effectiveness of this time-step and modality-aware mixing strategy. 

Overall, the main contribution of our paper can be summarized as follow:
\begin{itemize}
\item To the best of our knowledge, this is the first work to introduce a mixup-perceptive strategy into spiking knowledge transfer. 
We propose a novel Time-step Mixup framework that randomly replaces individual appearance frames with event frames at various time steps, which complemented by additional smoothed modality labels.

\item We introduce two auxiliary learning objectives, namely the Modality-Aware Guidance (MAG) label and the Mixup Ratio Perception (MRP) label, to assist the proposed Time-step Mixup. These are designed to enhance the model's capability in learning discriminative and temporally consistent representations across different modalities.
\end{itemize}

\section{Related Work}
\subsection{Spiking Neuron Models}
SNNs draw inspiration from the human brain, using discrete spikes for information processing. This method achieves effects comparable to continuous activation functions by accumulating spikes over an additional temporal dimension, making it highly suitable for processing temporal data. Concretely, SNNs replace the traditional activation function by using a spiking neuron model, such as the Integrate-and-Fire (IF) neuron model \cite{IF}  and the widely-used Leaky Integrate-and-Fire (LIF) neuron model \cite{LIF}. The LIF neuron model integrates incoming spikes over time, with its membrane potential and spiking behavior governed by the following equations:
\begin{equation}
    \textbf{u}^{t+1,l} = \tau \textbf{u}^{t,l} + \textbf{W}^{l} \textbf{s}^{t,l-1}
\end{equation}
\begin{equation}
    \textbf{s}^{t,l} = H(\textbf{u}^{t,l} - V_{\text{th}})
    \label{eq2}
\end{equation}
\begin{equation}
    \textbf{u}^{t+1,l} = \tau \textbf{u}^{t,l} \cdot (1 - \textbf{s}^{t,l}) + \textbf{W}^{l} \textbf{s}^{t+1,l-1}
\end{equation}
where \( \textbf{u}^{t,l} \) denotes the membrane potential of neurons in layer \( l \) at time-step \( t \), \( \textbf{W}^{l} \) represents the weight matrix of layer \( l \), and \( \textbf{s}^{t,l} \) corresponds to the binary spikes emitted by neurons. The Heaviside step function \( H \) determines whether a spike is emitted, based on the comparison between \( \textbf{u}^{t,l} \) and the threshold \( V_{\text{th}} \). The leaky factor \( \tau \) controls the temporal decay of the membrane potential. 

\subsection{Spiking Knowledge Transfer}
Knowledge transfer has been widely applied in traditional ANNs and has achieved remarkable success \cite{wang2019domain}. 
However, in Spiking Neural Networks (SNNs), which have attracted increasing attention due to their energy efficiency, transfer learning remains relatively underexplored.
In particular, transferring knowledge from appearance domains (e.g., grayscale or color images) to the event domain (e.g., DVS data) using SNNs holds great potential for addressing the challenges of limited scale and accessibility of DVS datasets, which often lead to poor generalization performance.
Although research in this area remains limited, a few recent works have started to explore knowledge transfer from static to event domains.

Specifically, R2ETL~\cite{stl} utilizes labeled RGB data for SNN transfer learning by introducing encoding and feature alignment modules, and extends CKA to TCKA. EKT~\cite{ekt} proposes a gradual replacement strategy, where static images are progressively replaced by event data during training, guided by a loss combining domain alignment and spatio-temporal regularization. CKD~\cite{ckd} adopts phased cross-architecture distillation, transferring appearance-domain features from ANNs to SNNs. 
These methods rely on shared-parameter backbones but overlook distribution gaps between source and input domains, leading to suboptimal transfer. 
Our paper proposes a smoother transfer approach, with a modality-aware guidance that mitigates the issues caused by input data distribution differences.

\subsection{Mixup for Cross-Modality Transfer}
Many works~\cite{guo2019mixup,hu2021neural,wang2022vlmixer} leverage MixUp-style data augmentation to bridge modality gaps. AdaMixUp \cite{guo2019mixup} views MixUp as a form of out-of-manifold regularization and addresses its limitations via adaptive mixing strategies. Neural Dubber \cite{hu2021neural} introduces a multi-modal TTS system that synchronizes speech with video using lip movements and speaker embeddings. VLMixer \cite{wang2022vlmixer} applies cross-modal CutMix for unpaired vision-language pretraining, achieving effective alignment between image and text modalities. Despite their success in other domains, MixUp-style strategies remain unexplored for transfer learning in SNNs. Transferring from appearance to event domains is particularly challenging due to large intensity distribution gaps, which make direct interpolation ambiguous. Notably, SNNs process static inputs by repeating frames across time-step and averaging outputs. Leveraging this structure, we propose a temporal replacement strategy that substitutes entire frames along the time-step axis, preserving semantics while enabling smoother cross-domain transfer.

\section{Methodology}
In this section, we present a new SNN-based framework, namely Time-step Mixup Knowledge Transfer (TMKT) model, designed to transfer knowledge from the appearance domain to the event domain. 
Specifically, TMKT integrates three key components: the Time-step Mixup strategy (TSM), the Modality-Aware Guidance module (MAG), and the Mixup ratio Perception module (MRP).
By leveraging the inherent temporal heterogeneity of spiking neural networks, TMKT constructs a robust common representation space between the source and target domains. This facilitates bridging the modality-induced domain gap and enables smooth and efficient transfer knowledge of source domain to the target domain.

\subsection{Overall Architecture}
\begin{figure*}[!t]
\centering
\includegraphics[width=0.95\textwidth]{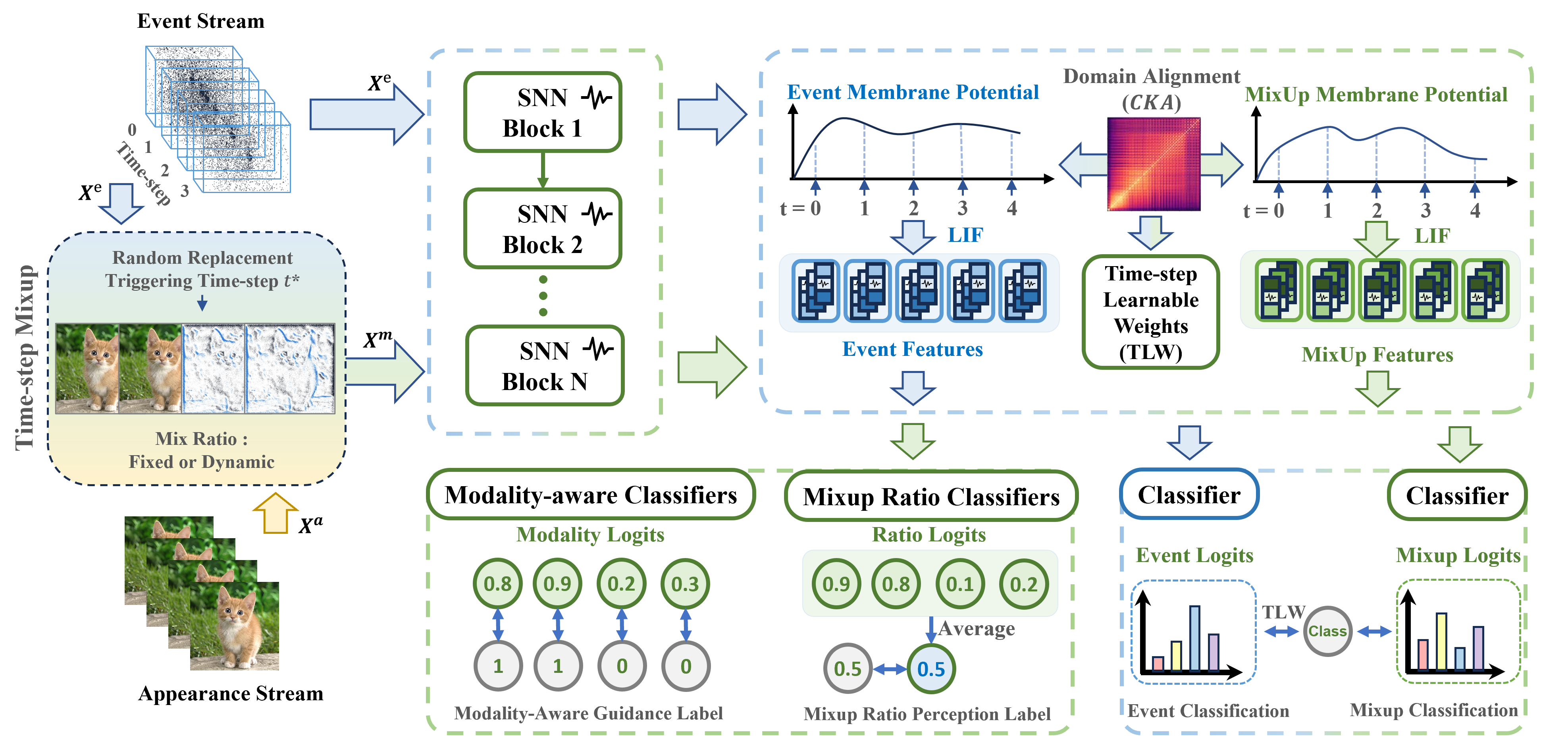} 
\caption{The overview of our proposed Time-step Mixup Knowledge Transfer (TMKT) framework. TMKT employs a Time-step Mixup (TSM) strategy and introduces two auxiliary labels: a modality-aware guidance label and a mixup ratio label to enhance the supervision of temporal knowledge transfer. 
Both the event stream and the Time-step Mixup stream are fed into the network simultaneously, sharing all weights except for the final layer. Membrane potentials from the penultimate layer are used for domain alignment.}
\label{main_fig}
\end{figure*}

As shown in Fig.~\ref{main_fig}, our TMKT model adopts a two-stream input paradigm, where paired appearance and event streams $\mathbf{X}^{\mathrm{a}}$ and $\mathbf{X}^{\mathrm{e}}$ of the same category are provided as input. 
These sequences are first processed by the TSM module to construct time-specific mixed ones $\mathbf{X}^{\mathrm{m}}$, where the modality components are interleaved across time-step.
The resulting sequence $\mathbf{X}^{\mathrm{m}}$ is then forwarded to a spiking neural network (SNN)-based backbone for feature extraction. 
During this stage, we introduce a Regularized Domain Alignment loss $\mathcal{L}_{\mathrm{RDA}}$ to align the feature distributions of mixed and event modalities within the mixed representation space, mitigating cross-domain discrepancies at the feature level.

To cooperate with the time-step mixed data for effective knowledge transfer, we introduce two novel modality-aware objectives at local and global levels respectively. At each time-step, a Modality-aware Guidance (MAG) loss $\mathcal{L}_{\mathrm{MAG}}$ is crafted to encourage the model to distinguish the dominant modality, promoting temporal consistency across streams. Meanwhile, another Mixup ratio Perception (MRP) loss $\mathcal{L}_{\mathrm{MRP}}$ is proposed to offer global supervision by estimating
the underlying mixing ratio applied by TSM.

\subsection{Time-step Mixup Strategy}
Before we dive into the details of Time-step Mixup, let us recall the general pipeline of SNNs again. The the appearance or event input are usually in the form frame sequences, denoted as $\mathbf{X}^{\mathrm{a}}=\left\{ \mathbf{x}^{\mathrm{a}}_{1}, \mathbf{x}^{\mathrm{a}}_{2}, \ldots, \mathbf{x}^{\mathrm{a}}_{T} \right\}$ and $\mathbf{X}^{\mathrm{e}}=\left\{ \mathbf{x}^{\mathrm{e}}_{1}, \mathbf{x}^{\mathrm{e}}_{2}, \ldots, \mathbf{x}^{\mathrm{e}}_{T} \right\}$ respectively. \( T \) is the total number of discrete Time-step.
If passing  $\mathbf{X}^{\mathrm{e}}$ through $N$ SNN blocks, corresponding event features $\mathbf{F}^\mathrm{e}_t$ at each step can be obtain as,
\begin{equation}
\mathbf{F}^\mathrm{e}_t = Enc(\mathbf{x}^{\mathrm{e}}_{t}), \quad t \in [1, T]
\label{eq5}
\end{equation}
where $Enc(\cdot)$ denotes the SNN feature extractor. A classifier \( \mathcal{F}(\cdot) \) is followed to obtain temporal logits as,
\begin{equation}
\mathbf{y}^{\mathrm{e}}_t = \mathcal{F}(\mathbf{F}^{\mathrm{e}}_{t}). \quad t \in [1, T]
\end{equation}
The temporal efficient training (TET) \cite{tet} loss can be used to integrate Cross-Entropy loss at each Time-step as,
\begin{equation}
\mathcal{L}_{\mathrm{TET}} = \frac{1}{T} \sum_{t=1}^{T} CE(\mathbf{y}^\mathrm{e}_t, \mathbf{y}),
\label{TET}
\end{equation}
where \( \mathbf{y} \) is the ground truth label and \( CE(\cdot) \) denotes the cross-entropy loss. 
From above equations, it can be seen that SNNs are naturally trained with sequential data containing multiple Time-step. Thus, our Time-step Mixup (TSM) method starts from randomly replacing individual appearance frames with event ones at different Time-step. 

Noting that, to avoid unstable generalization caused by frequent modality switching across Time-step, we adopt a truncation replacement strategy.
Given an expected Mixup ratio $r_m$, i.e. the portion of replaced samples, we have a corresponding replacement probability $p$ at every time-step, which will be calculated later.
For each appearance sample, starting from the first time-step $t=1$, we sequentially sample a uniform random variable $u_t \sim \mathcal{U}(0,1)$ at each time-step $t$. If $u_t < p$, we trigger replacement at this frame and substitute all subsequent frames with event-domain data. Formally, the replacement point $t^*$ is determined as,
\begin{equation}
t^* = \min \left\{ t \in \{1, 2, \ldots, T\} \,\middle|\, u_t < p \right\}
 \end{equation}
where it follows a truncated geometric distribution. Given the time-step length $T$, the probability of having $t^*=1,2,\ldots,T$ is calculated as,
\begin{equation}
P(X=t^* \mid X \leq T) = \frac{(1-p)^{t^*-1}p}{1-(1-p)^{T}}
\end{equation}
Then the expectation of the replaced frame number can be computed and it shall align with the Mixup ratio $r_m$ as,
\begin{equation}
\begin{aligned}
\mathbb{E}[X] &= \sum_{t=1}^{T}(T+1-t)\cdot\frac{(1-p)^{t-1}p}{1-(1-p)^{T}} \\
              &= T \cdot {r}_{m}
\end{aligned}
\label{eq:expect}
\end{equation}
Given the values of $T$ and ${r}_{m}$, the replacement probability $p$ can be obtain by solving Eq. \ref{eq:expect}. Unfortunately, it has no closed-form solution, thus we approximate $p$ using numerical methods.
 
The final mixed input sequence $\mathbf{X}^{\mathrm{m}}=\left\{ \mathbf{x}^{\mathrm{m}}_{1}, \mathbf{x}^{\mathrm{m}}_{2}, \ldots, \mathbf{x}^{\mathrm{m}}_{T} \right\}$ is then constructed as,
 \begin{equation}
 \mathbf{x}_t^{\mathrm{m}} =
 \begin{cases}
 \mathbf{x}_t^\mathrm{a}, & \text{if } \ t < t^* \\
 \mathbf{x}_t^\mathrm{e}, & \text{if } \ t \geq t^*
 \end{cases}
 \label{eq:mixup}
 \end{equation}
If no replacement occurs, $t^* = T+1$, i.e. the sequence is identical to the original appearance input.

\subsection{Domain Alignment}
After obtaining the mixed data $\mathbf{X}^\textrm{m}$, we feed them and the target domain (event) input $\mathbf{X}^e$ into a parameter-shared model, to generate mixed features and target features respectively.
As shown in Eq.~\ref{eq2}, when the membrane potential is below the threshold, it will be retained in the neuron without triggering a spike. 
In particular, the membrane potentials remaining in the final layer encode rich, high-precision features and are informative for feature space alignment. 
Therefore, we collect the membrane potentials in the final layer's neurons of the SNN feature extractor $Enc(\cdot)$, given in Eq.\ref{eq5}, to obtain a robust common representation space.

Subsequently, we adopt and perform domain alignment to minimize the distance between the common space and the event feature space, ensuring effective knowledge transfer.
As illustrated in Fig.\ref{main_fig}, we employ CKA \cite{cka}, a widely used metric for its effectiveness in measuring network representation similarity, to align the two domains within the SNN. 
Therefore, we collect the remaining membrane potentials from the layer preceding the final layer. At \textit{t}-th time-step, the potentials for \textit{i}-th and \textit{j}-th categories in mixed and event streams are denoted as $\mathbf{V}^\mathrm{m}_{i,t}$ and $\mathbf{V}^\mathrm{e}_{j,t}$. The domain alignment loss $\mathcal{L}_{\mathrm{DA}}$ is then calculated based on CKA as,
\begin{equation}
\mathcal{L}_{\mathrm{DA}}=1-\frac{1}{T}\sum_{t=1}^T\mathop{CKA}_{\substack{\mathbf{y}_i=\mathbf{y}_j\\ \mathbf{y}\in{\mathcal{\mathbf{Y}}}}}\left(\mathbf{V}^\mathrm{m}_{i,t},\mathbf{V}^\mathrm{e}_{j,t}\right)
\label{eqi:domain alignment_1}
\end{equation}
where $\mathbf{y}$ is the class label of the input data, drawn from the overall class set $\mathcal{\mathbf{Y}}$. 
The condition $\mathbf{y}_i=\mathbf{y}_j$ indicates a matched pair of mixup and event data. 

In addition, we assign learnable weights to each time-step, and introduce an event data classification loss ${\mathcal{L}}_{CLS_e} = \mathcal{L}_\text{TET}$ as a regularization term to prevent the model from overfitting to specific time-step. Thus, the regularized domain alignment loss $\mathcal{L}_{\mathrm{RDA}}$ is defined as,
\begin{equation}
\begin{aligned}
\mathcal{L}_{\mathrm{RDA}} =\frac{1}{T}\sum_{t=1}^{T} & \bigg(\sigma(\theta_{t}) \big( 1 - \mathop{CKA}_{\substack{\mathbf{y}_i=\mathbf{y}_j\\ \mathbf{y}\in{\mathcal{\mathbf{Y}}}}}\left(\mathbf{V}^\mathrm{m}_{i,t}, \mathbf{V}^\mathrm{e}_{j,t}\right) \big) \\
& \quad + \big(1 - \sigma(\theta_{t})\big) \cdot \mathcal{L}_{\mathrm{CLS_e}} \bigg),
\end{aligned}
\label{equ:domain_alignment}
\end{equation}
where $\sigma(\cdot)$ denotes the sigmoid function, and $\theta_{t} $ is the learnable weight for time-step $t$. 

\subsection{Time-step Mixup Labels}
As part of our innovative approach to the Mixup strategy, we introduce a modality-aware guidance module and a mixup ratio perception module as shown in the green dashed box in Fig. \ref{main_fig}. These components offer detailed cues essential for learning modality-specific information across various time-steps. When engaging in cross-modal mixing, it's crucial for the model to differentiate between appearance and event inputs to prevent confusion arising from their distinct distributions.By integrating these modules with input data Mixup, we achieve a smoothing effect on both feature and label distributions similar to traditional Mixup methods, which subsequently enhances the generalization capabilities of Spiking Neural Network (SNN) models.To further facilitate this process, we've designed two types of labels aimed at explicitly guiding the model in understanding modality characteristics and mastering mixing patterns:

\textbf{Frame-wise Modality-Aware Label} indicates the source (appearance / event) of each frame at the corresponding time-step, addressing the ambiguity of mixed input sources. For a mixed sample $\mathbf{X}^{\text{m}}$, the modality-aware label of the $t$-th frame, $y_t^s \in \{0, 1\}$, is defined as,
\begin{equation}
y^\mathrm{s}_t =
\begin{cases} 
0, & \text{if } \mathbf{x}_t^\mathrm{m} \in \mathbf{X}^\mathrm{e} \\ 
1, & \text{if } \mathbf{x}_t^\mathrm{m} \in \mathbf{X}^\mathrm{a}
\end{cases}
\label{eq:modality label}
\end{equation}

This label acts as a modality signal that guides the model to recognize the intrinsic differences between appearance and event features, preventing misalignment of their representations.

\textbf{Sample-wise Mixup ratio Perception Label} reflects the proportion of appearance frames in the mixed sample, quantifying the mixing degree to guide the model in understanding the transition pattern. For a mixed sample of length $T$, let the number of appearance frames be \(K = t^* - 1\). The Mixup ratio perception label \(y_m \in [0, 1]\) is defined as,
\begin{equation}
y_{m}\,=\,{\frac{K}{T}}
\label{eq:mixup label}
\end{equation}

This label helps the model learn the temporal structure of cross-modal mixing, reinforcing the connection between input mixing patterns and semantic representations.

\subsection{Loss Function}
To ensure the model effectively learns from mixed data and modality cues, we design a dual-loss framework that combines modality discrimination and mixing ratio estimation:

\textbf{Modality-Aware Guidance Loss} enforces the model to accurately identify the source of each frame, enhancing its ability to distinguish appearance and event distributions. Let the SNN predict the source of the $t$-th frame as $\hat{\mathbf{z}}^s_t = g_s(\mathbf{F}_t)$, where $\mathbf{F}_t$ is the SNN's spike feature at frame $t$, and $g_s(\cdot)$ is the Modality-aware classification head. The Modality-aware Guidance loss adopts the cross-entropy loss as,
\begin{equation}
{\mathcal{L}}_\mathrm{MAG}={\frac{1}{T}}\sum_{t=1}^{T}CE({\hat{\mathbf{z}}}^{s}_{t},{y}^{s}_{t})
\label{eq:modality loss}
\end{equation}

\textbf{Mixup ratio Perception Loss} constrains the model to estimate the overall mixing ratio of the sample, encouraging it to capture the global temporal transition pattern. Let the SNN predict the Mixup ratio as \(\hat{z}_m = \frac{1}{T} \sum_{t=1}^T g_m(\mathbf{F}_t)\), where \(g_m(\cdot)\) is the Mixup ratio prediction head. The Mixup ratio perception loss uses mean squared error as,
\begin{equation}
{\mathcal{L}}_\mathrm{MRP}=\operatorname{MSE}({\hat{z}}_{m},y_{m})
\label{eq:mixup loss}
\end{equation}

Finally, the total classification loss is defined as,
\begin{equation}
\mathcal{L}_{\mathrm{total}} = \mathcal{L}_{\mathrm{CLS_m}} +\lambda\mathcal{L}_{\mathrm{RDA}} + \mathcal{L}_{\mathrm{MAG}} + \mathcal{L}_{\mathrm{MRP}}
\label{eq:total loss}
\end{equation}
where $\mathcal{L}_{\mathrm{CLS_m}}$ denotes the classification loss on Mixup data, and $\lambda$ is the weighting coefficient for $\mathcal{L}_{\mathrm{RDA}}$.

\section{Experiments}
Our experiments are conducted on several mainstream event-based datasets, including N-Caltech101 \cite{ncal}, and N-Omniglot \cite{Nomniglot}, along with their corresponding RGB-based counterparts. We also conduct experiments on CEP-DVS \cite{cepdvs}, an image-event paired dataset.
\begin{table*}[!t]
\centering

\begin{tabular}{cccccc}
\hline
\textbf{Dataset}                & \textbf{Category}                    & \textbf{Methods}                              & \textbf{Architecture}             & T                 & \textbf{Accuracy}                      \\ \hline
                                &                                      & NDA \cite{nda}                           & VGGSNN                            & 10                         & 78.2                                   \\
                                & \multirow{-2}{*}{Data augmentation}  & EventMixer \cite{eventmix}            & ResNet-18                         & 10                         & 79.5                                   \\ \cdashline{2-6} 
                                &                                      & TET \cite{tet}                          & VGGSNN                            & 10                         & 79.27                                  \\
                                &                                      & TCJA-TET \cite{tcja}                      & CombinedSNN                       & 14                         & 82.5                                   \\
                                &                                      & TKS \cite{tks}                    & VGGSNN                            & 10                         & 84.1                                   \\
                                & \multirow{-4}{*}{Efficient training} & ETC \cite{etc}                          & VGGSNN                            & 10                         & 85.53                                  \\ \cdashline{2-6} 
                                &                                      & R2ETL with TCKA \cite{stl}              & VGGSNN                            & 10                         & 82.70                                  \\
                                &                                      & Knowledge-Transfer \cite{ekt}            & VGGSNN                            & 10                         & 93.18                                  \\
                                &                                      & CKD \cite{ckd}                                      & VGGSNN                            & 10                         & \underline{97.13}                                  \\
\multirow{-10}{*}{N-Caltech101} & \multirow{-4}{*}{Transfer learning}  & \cellcolor[HTML]{EFEFEF}TMKT (Ours) & \cellcolor[HTML]{EFEFEF}VGGSNN    & \cellcolor[HTML]{EFEFEF}10 & \cellcolor[HTML]{EFEFEF}\textbf{97.93} \\ \hline
                                & Efficient training                   & TET \cite{tet}                          & ResNet-18                         & 6                          & 25.05                                  \\ \cdashline{2-6} 
                                & Data extension                       & Ev2Vid \cite{ev2vid}                                       & ResNet-18                         & 6                          & \underline{31.20}                                  \\ \cdashline{2-6} 
                                &                                      & Knowledge-Transfer \cite{ekt}            & ResNet-18                         & 6                          & 30.50                                  \\
\multirow{-4}{*}{CEP-DVS}       & \multirow{-2}{*}{Transfer learning}  & \cellcolor[HTML]{EFEFEF}TMKT (Ours) & \cellcolor[HTML]{EFEFEF}ResNet-18 & \cellcolor[HTML]{EFEFEF}6  & \cellcolor[HTML]{EFEFEF}\textbf{34.70}  \\ \hline
                                & Efficient training                   & plain \cite{Nomniglot}                         & SCNN                              & 12                         & 60.0                                  \\ \cdashline{2-6} 
                                &                                      & Knowledge-Transfer \cite{ekt}            & SCNN                              & 12                         & \textbf{63.60}                                  \\
\multirow{-3}{*}{N-Omniglot}    & \multirow{-2}{*}{Transfer learning}  & \cellcolor[HTML]{EFEFEF}TMKT (Ours) & \cellcolor[HTML]{EFEFEF}SCNN      & \cellcolor[HTML]{EFEFEF}12 & \cellcolor[HTML]{EFEFEF}\underline{63.09}             \\ \hline
\end{tabular}

\caption{Comparison between the proposed method and existing works. Bold and underline items indicate the best andsecond-best results, respectively.}
\label{table1}

\end{table*}

\begin{table}[t]
\centering

\begin{tabular}{ccccc}
\hline
\multicolumn{1}{c|}{Network}           & TSM & $\mathcal{L}_\mathrm{MAG}$ & \multicolumn{1}{c|}{$\mathcal{L}_\mathrm{MRP}$} & Accurary \\ \hline
\multicolumn{5}{c}{\textbf{N-Caltech101}}                                                                                                                   \\ \hline
\multicolumn{1}{c|}{\multirow{5}{*}{VGGSNN}}    & -             & -                        & \multicolumn{1}{c}{-}                        & 93.45             \\
\multicolumn{1}{c|}{}                           & \checkmark    & -                        & \multicolumn{1}{c|}{-}                        & 97.24             \\
\multicolumn{1}{c|}{}                           & \checkmark    & \checkmark               & \multicolumn{1}{c|}{-}                        & 97.36             \\
\multicolumn{1}{c|}{}                           & \checkmark    & -                        & \multicolumn{1}{c|}{\checkmark}               & 97.70             \\
\multicolumn{1}{c|}{}                           & \checkmark    & \checkmark               & \multicolumn{1}{c|}{\checkmark}               & \textbf{97.93}    \\ \hline
\multicolumn{5}{c}{\textbf{CEP-DVS}}                                                                                                                        \\ \hline
\multicolumn{1}{c|}{\multirow{5}{*}{ResNet-18}} & -             & -                        & \multicolumn{1}{c|}{-}                        & 30.50             \\
\multicolumn{1}{c|}{}                           & \checkmark    & -                        & \multicolumn{1}{c|}{-}                        & 33.00             \\
\multicolumn{1}{c|}{}                           & \checkmark    & \checkmark               & \multicolumn{1}{c|}{-}                        & 32.80             \\
\multicolumn{1}{c|}{}                           & \checkmark    &  -                       & \multicolumn{1}{c|}{\checkmark}               & 33.55             \\
\multicolumn{1}{c|}{}                           & \checkmark    & \checkmark               & \multicolumn{1}{c|}{\checkmark}               & \textbf{34.70}    \\ \hline
\end{tabular}

\caption{Ablation experiments of Time-step Mixup Knowledge Transfer. TSM refers to Time-step Mixup, $\mathcal{L}_\mathrm{MAG}$ refers to Modality-aware Guidance Loss, $\mathcal{L}_\mathrm{MRP}$ refers to Mixup ratio Perception Loss.}
\label{table2}
\end{table}

\begin{table}[t]
\centering

\begin{tabular}{cccc}
\hline
\textbf{Network}                 & \textbf{Dataset}                       & \textbf{Mixup ratio} & \textbf{Accurary}       \\ \hline
\multirow{5}{*}{VGGSNN} & \multirow{5}{*}{N-Caltech101} & 0.3         & 97.36           \\
                        &                               & 0.4         & \textbf{97.93} \\
                        &                               & 0.5         & 97.59          \\
                        &                               & 0.6         & 97.47            \\
                        &                               & 0.7         & 97.70     \\
\hline
\end{tabular}

\caption{Ablation experiments of Time-step Mixup ratio ${r}_{m}$.}
\label{table3}
\end{table}

\begin{table}[t]
\centering

\begin{tabular}{ccc}
\hline
\multicolumn{3}{c}{\textbf{N-Caltech101}}                            \\ \hline
\textbf{Network}        & \textbf{Mixup strategy} & \textbf{Accuary} \\ \hline
\multirow{4}{*}{VGGSNN} & Fixed Ratio             &  95.86                \\
                        & Dynamic Ratio (Non-Linear) &  95.05           \\
                        & Dynamic Ratio (Linear)  &  96.55           \\                        
                        & Time-step Mixup         & \textbf{97.93}    \\ \hline
\end{tabular}

\caption{Ablation experiments of the Mixup Strategy.}
\label{table4}
\end{table}

\subsection{Experimental Settings}

For a fair comparison, we follow the implementation of our baseline \cite{ekt}, which set the input size of N-Caltech101, CEP-DVS and N-Omniglot to 48, 48 and 28, as well as their RGB counterparts.
Model-wise, we use VGGSNN (10 time-steps, 300 epochs) for N-Caltech101, ResNet18 (6 time-steps, 200 epochs) for CEP-DVS, and SCNN (12 time-steps, 50 epochs) for N-Omniglot.
Regarding input encoding, static images are directly encoded and transformed into the HSV (Hue, Saturation, Value) color space to minimize the mismatch between appearance and event data. Considering the dual-channel characteristics of event data (positive and negative polarities), we replicate the value channel of the HSV representation and duplicate the static image across time-steps uniformly.
For the Mixup ratio ${r}_{m}$, we set it to 0.4, 0.9 and 0.95 on these datasets, respectively. 
The coefficient $\lambda$ of the loss given in Eq.~\ref{eq:total loss} is set to 0.5 in all experiments. All experiments are implemented based on the BrainCog framework \cite{braincog}.

\subsection{Comparison with the State-of-the-Art}
The experimental results on N-Caltech101, CEP-DVS, and N-Omniglot are summarized in Tab.~\ref{table1} for detailed comparison. The method categorization follows the baseline \cite{ekt}.
On N-Caltech101, our method is compared with several recent state-of-the-art approaches, including three different categories (data augmentation, efficient learning and transfer learning).
Under the VGGSNN architecture, our method achieves a new state-of-the-art accuracy of 97.93\%, demonstrating the superior effectiveness of the proposed TMKT framework. It outperforms our baseline method Knowledge-Transfer \cite{ekt} by a notable margin of 4.75\%.  On CEP-DVS, we also observe a constant performance gain, achieving a 4.2\% improvement over the baseline.

N-Omniglot is a few-shot event-based dataset characterized by a limited number of samples per class, which is challenging due to its outdated collection protocols and inherent noise/artifacts in the released version. Our method achieves an accuracy of 63.09\%, which is slightly lower than the baseline-reported result. However, it is important to note that under strictly identical implementation settings, our re-implementation of the baseline only reaches 60.69\%, indicating that our TMKT framework still outperforms the baseline by +2.4\% in a fair comparison.

\subsection{Ablation Study}
To verify the effectiveness of our method, extensive ablation studies are conducted comparing to our baseline Knowledge-Transfer \cite{ekt}.
\begin{figure*}[!t]
\centering
    \begin{subfigure}{0.32\linewidth}
        \centering
        \includegraphics[width=\linewidth]{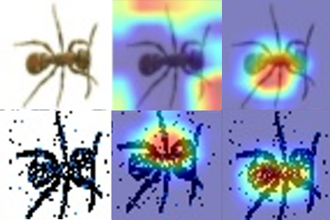}
        \caption{ant}
    \end{subfigure}
    \hfill
    \begin{subfigure}{0.32\linewidth}
        \centering
        \includegraphics[width=\linewidth]{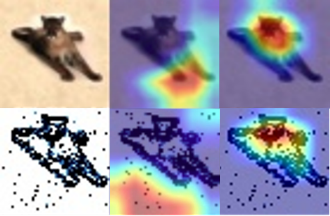}
        \caption{cougar body}
    \end{subfigure}
    \hfill
    \begin{subfigure}{0.32\linewidth}
        \centering
        \includegraphics[width=\linewidth]{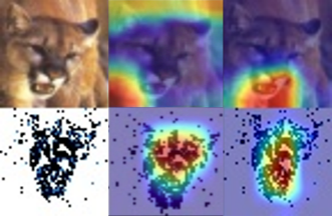}
        \caption{cougar face}
    \end{subfigure}
\caption{Class Activation Mapping of Caltech101 and N-Caltech101. For each class, the top row shows static images, and the bottom row presents event data integrated into frames. Within each class, from left to right are: original input, baseline result, and the result of our method.}
\label{fig:cam}
\end{figure*}

\begin{figure}[!t]
\centering
    \begin{subfigure}{0.48\linewidth}
        \centering
        \includegraphics[width=\linewidth]{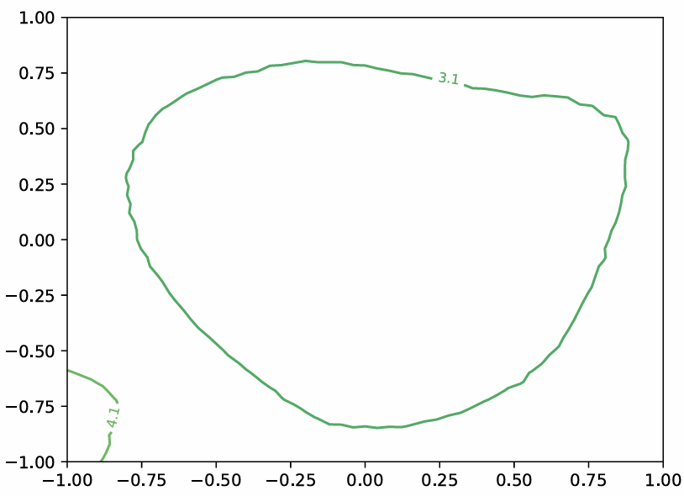}
        \caption{Baseline}
    \end{subfigure}
    \hfill
    \begin{subfigure}{0.48\linewidth}
        \centering
        \includegraphics[width=\linewidth]{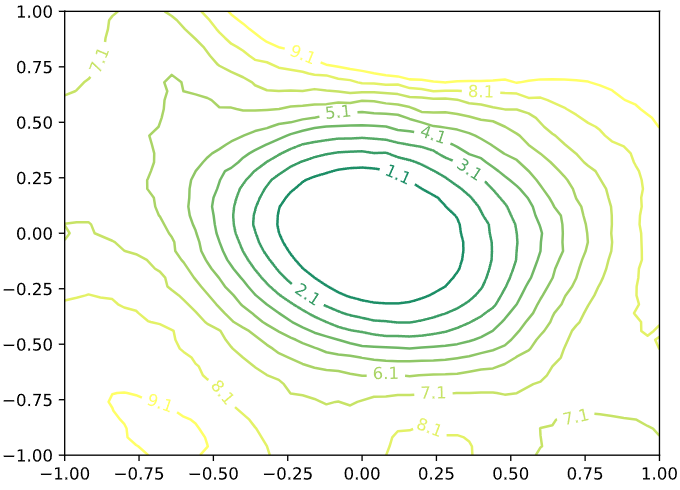}
        \caption{Ours}
    \end{subfigure}
\caption{Visualization of the loss landscapes for our method and the baseline on CEP-DVS dataset.}
\label{fig:loss_landscape}
\end{figure}

\subsubsection{Time-step Mixup Knowledge Transfer.}
We evaluated our proposed TMKT on both VGGSNN and ResNet-18, with the results summarized in Tab. \ref{table2}. 
Compared to the baseline, applying the Time-step Mixup alone yields a significant performance improvement, demonstrating its effectiveness in helping the model smoothly learn from a common representation space across modalities. 
When supervised by the two proposed auxiliary losses, one of which indicates the dominant modality at each timestep while the other encodes the overall mixup ratio of multimodal information for the entire sample, the model consistently achieves improved performance. When combined, these two losses lead to the best overall results.

\subsubsection{Mixup ratio.}
The mixup ratio \( r_{m} \) is a crucial hyperparameter in our framework, as it determines the overall mixing proportion between appearance data and event data in the Time-step Mixup process.  
We conduct ablation experiments on the N-Caltech101 dataset, as shown in Tab.~\ref{table3}, and find that \( r_{m} = 0.4 \) yields the best performance.  
Notably, all other ablated settings still significantly outperform the baseline Knowledge-Transfer \cite{ekt}, further demonstrating its effectiveness and robustnes.

\subsubsection{Mixup strategy.}
We further conduct an ablation experiment on various Mixup strategies for constructing Time-step Mixup data. As presented in Tab. \ref{table4}, the fixed ratio strategy refers to replace appearance parts with those of event data at a fixed mixup ratio. The $t^*$ in Eq.~\ref{eq:mixup} is set to a constant value $\lfloor T \cdot r_{\text{m}} \rfloor$. 
Another strategy is dynamic ratio, inspired by the baseline \cite{ekt}, adopting a progressive replacement scheme. 
Two configurations of dynamic ratio $r'_{m}$ and $r''_{m}$ are tested, namely linear and nonlinear ones, which are determined using the following functions,
\begin{align}
r_{m}'&=\left(({b_{i}+e_{c}*b_{l}})/({e_{m}*b_{l}})\right)^{3},\\
r_{m}''&=\left({e_{c}}/{e_{m}}\right), 
\end{align}
where $b_i$ denotes the current batch index, $b_l$ is the total number of batches per epoch, $e_c$ is the current epoch number, and $e_m$ represents the total number of training epochs. Under this strategy, the appearance stream is gradually replaced by the event stream as training progresses. Compared to these strategies, our probabilistic strategy produces more diverse and flexible data, which help the model to learn a more robust and generalizable common representational space.

\subsection{Analysis and Discussion}
\subsubsection{Cross-Modal Visual Interpretability.}
To further assess whether our method successfully learns a common representational space across the appearance and event domains, we adopt grad-cam++ \cite{grad_cam} for visual explanation. This technique highlights the image regions that contribute most to the final classification decision. As shown in Fig. \ref{fig:cam}, our method consistently focuses on the key semantic regions in both appearance and event inputs, demonstrating its ability to bridge modality differences and extract shared discriminative features.

\subsubsection{Loss Landscape.}
To investigate whether our method enables the SNN to learn more discriminative features in the event domain, we perform experiments using 2D loss landscape visualization \cite{losslandscape} on the CEP-DVS dataset, comparing our method with the baseline.  
As demonstrated in Fig.~\ref{fig:loss_landscape}, our approach produces a more compact and concentrated loss basin around the minimum, suggesting that the model converges to a sharper and more well-defined solution.  
This indicates that our training strategy facilitates the learning of more discriminative representations within the event domain.  
In contrast, the baseline exhibits a flatter and more irregular surface. It may reflect a less stable convergence behavior and potentially inferior generalization capability.

\section{Conclusion}
In this paper, we have proposed a Time-step Mixup Knowledge Transfer framework for spiking neural networks to facilitate knowledge transfer from the appearance domain to the event domain. By mixing appearance and event sequences at the time-step level, and incorporating Modality-aware Guidance Loss, Mixup ratio Perception Loss and Domain Alignment, the framework has achieved smooth and effective cross-modal knowledge transfer. Experiments on N-Caltech101, CEP-DVS, and other datasets have demonstrated superior performance, establishing a novel paradigm for knowledge transfer in spiking neural networks. The future work will focus on more different and dedicated mixing strategies over both spatial and temporal domains.

\bigskip

\bibliography{aaai2026}

\end{document}